# Classifying a specific image region using convolutional nets with an ROI mask as input

Sagi Eppel[1]

## Abstract

Convolutional neural nets (CNN) are the leading computer vision method for classifying images. In some cases, it is desirable to classify only a specific region of the image that corresponds to a certain object. Hence, assuming that the region of the object in the image is known in advance and is given as a binary region of interest (ROI) mask, the goal is to classify the object in this region using a convolutional neural net. This goal is achieved using a standard image classification net with the addition of a side branch, which converts the ROI mask into an attention map. This map is then combined with the image classification net. This allows the net to focus the attention on the object region while still extracting contextual cues from the background. This approach was evaluated using the COCO object dataset and the OpenSurfaces materials dataset. In both cases, it gave superior results to methods that completely ignore the background region. In addition, it was found that combining the attention map at the first layer of the net gave better results than combining it at higher layers of the net. The advantages of this method are most apparent in the classification of small regions which demands a great deal of contextual information from the background.

## 1. Introduction

Convolutional neural nets (CNN) are the leading computer vision method for classifying objects in images[1]. In some cases, it is desirable to classify only a specific object in a given region of the image[2,3] (Figure 1).

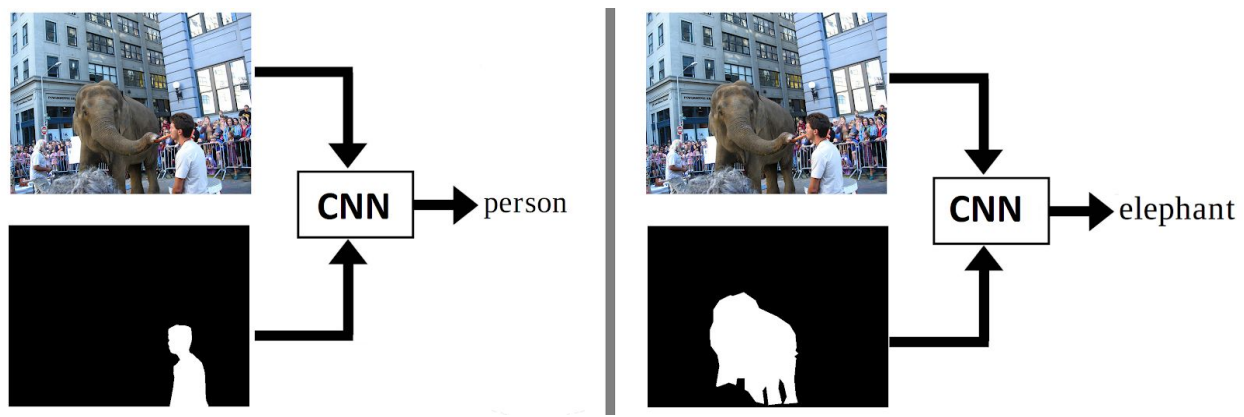

Figure 1. Region-specific image classification using a CNN. The region for classification is given as a region of interest (ROI) mask and used as an additional input to the net.

[1] sagieppel@gmail.com, Vayavision

Assuming that the segment of the object in the image is known and is given as a binary region of interest (ROI) mask, the goal is to classify the object in the region. One simple approach for achieving this is to black out the background area around the object segment[3,4,] (Figure 2a); another approach is to crop the object region and use it as a separate image for classification[5] (Figure 2b). Although both of these approaches were successfully applied[3-5] they suffer from a loss of background information, which can give significant contextual cues for identifying the object (Figure 2a). This information can be important for classification of small or blurred objects.

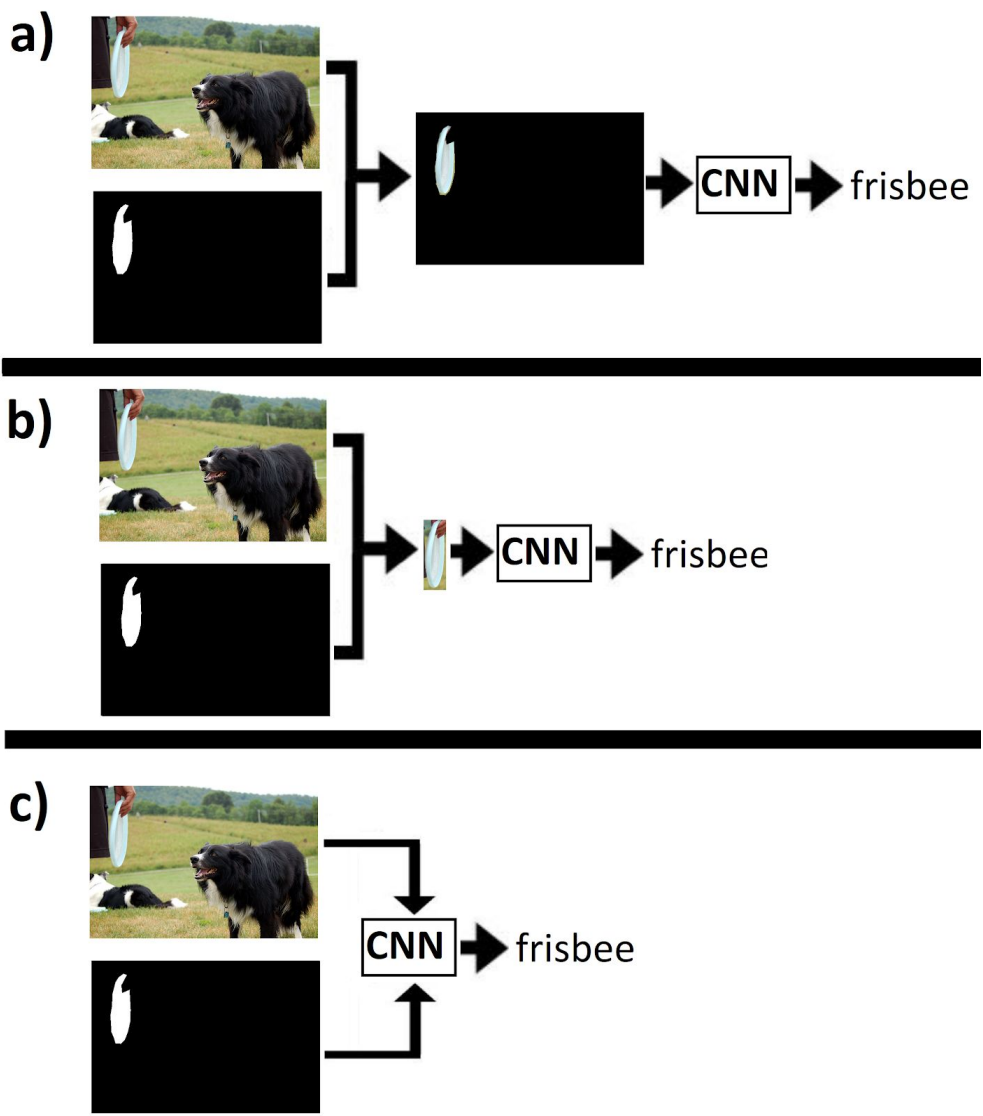

**Figure 2. Three different methods for using the attention region with neural nets: a) blacking out the background region; b) cropping the region around the ROI and using it as a separate image; c) using the ROI mask as an additional input to the net.**

An alternative approach is to generate an attention map, which can be used by the net to extract features from both objects and the background[3,6-8] (Figures 2c, 3). This approach involves using the ROI mask as an additional input to the net. The ROI mask is processed using a side branch of the net to generate an attention map, which is then combined with the main image classification branch (Figure 3). This allows the net to focus its attention on the object region of the image while still extracting contextual cues from the background region. An attention-based approach has been used in various areas, such as image-captioning[6,7], ROI classification[3], and hierarchical semantic segmentation[8], and has proven to be more effective than methods that completely ignore the background region. An attention map can easily be generated from the input ROI mask using a convolution layer (Figure 3). The main question, in this case, is how this map can be combined with the main image classification net to enable it to best focus attention on the ROI region while still using the background information.

## 2. Net architecture

Several net architectures for the classification of a specific image region were examined, and are shown in Figure 3. In general, all of these are based on the main branch, which consists of a Resnet50 image classification net for processing the input image[1] (Figure 3a), and a side branch that processes the ROI map, using a single convolution layer to generate an attention map. This attention map is then combined with one or more layers of the main branch, either by element-wise addition or multiplication (Figure 3b–g). The combined layer is then used as an input for the next layer of the main branch (Figure 3b–g). In order to allow element-wise addition or multiplication, the attention map must be the same size as the layer with which it is combined. To achieve this, the ROI mask was first resized to match the size of the layer with which it was merged, and a convolution layer was then applied (Figure 3d–e) with the same number of filters as the depth of the target layer. For cases where the attention maps were combined with more than one layer (Figure 3f–g), a separate attention map was generated using different convolution filters for each layer. Two methods for completely removing background information (Hard attention) were also examined by zeroing out all regions outside the ROI in the image (method 1) or in the feature map of the first layer (method 2).

### 2.1 Net initiation

The Resnet50 main branch was initialized using a model trained on ImageNet for full image classification. Only the last layer was changed, to fit the number of categories of the specific classification task. The convolution layer of the side branch was initialized as follows: if the attention map was to be merged by element-wise addition, both the weights and the bias were initialized to zero; if the attention map was to be merged multiplication, the bias was set to one and the filter weights to zero. This weights initiation method promise that the initial effect of the attention branch on the classification branch is zero at the outset and increases gradually during training.

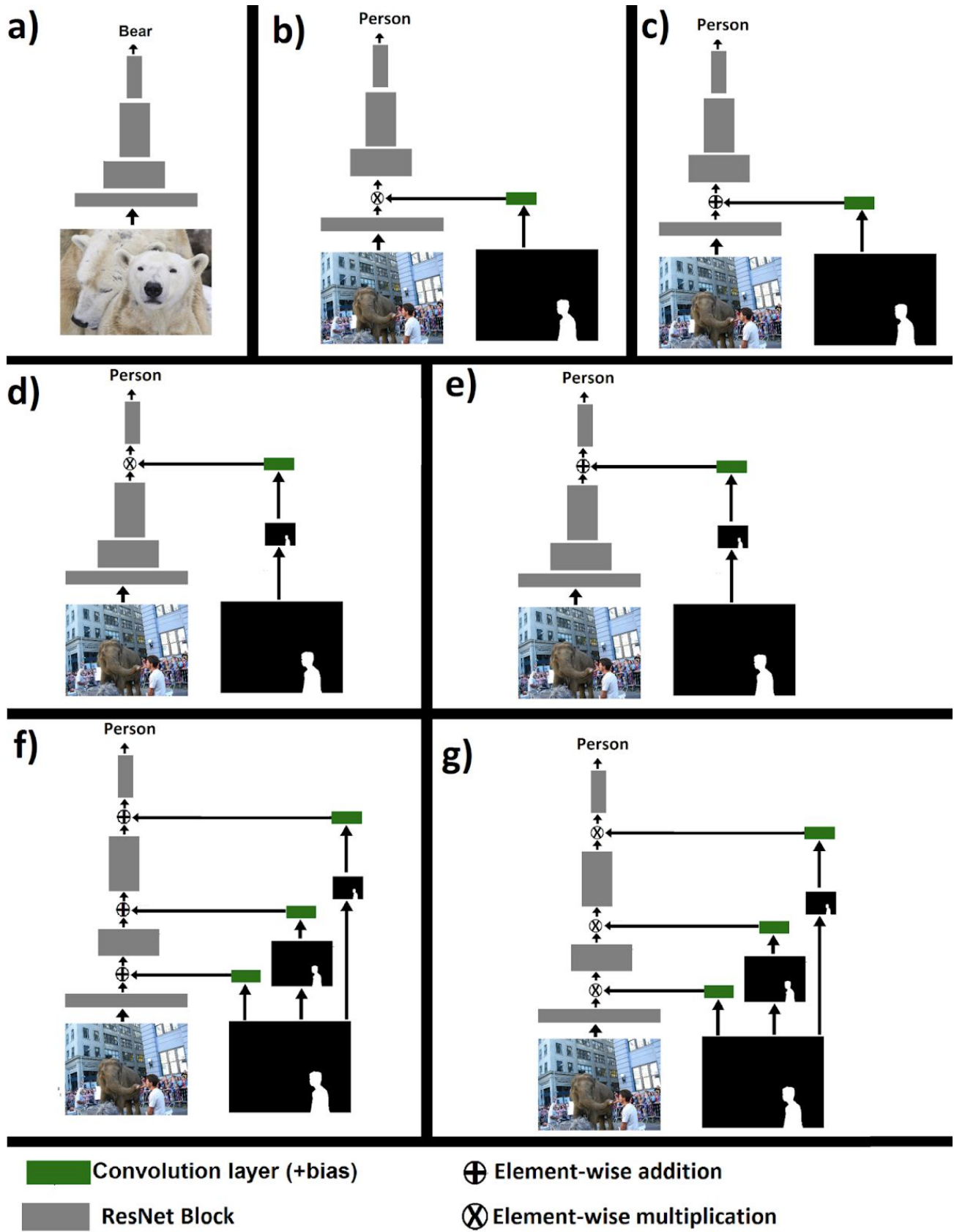

Figure 3. a) Resnet 50 classification of the full image; b-g) different approaches for combining the attention maps with the net.

## 3. Datasets

The methods were evaluated using the COCO object dataset[9]. The input ROI for the net was the object instance mask and the output was the object class. The nets were also trained using the OpenSurfaces material classification dataset[10]; in this case, the ROI was generated by taking a connected region of the image corresponding to a single material, and the output was the material type. The code and weights of the trained nets for both datasets are supplied in the supporting material section.

## 4. Evaluation

The evaluation was carried out by averaging the classification accuracy over all classes. Tables 1 and 2 give the mean accuracy average over all the classes. Hence, the mean classification accuracy was taken for each class separately and the result was averaged (Each class contributes equally to the statistics). Tables 3 and 4 give the mean accuracy over all images. Hence, more common classes contribute more to the mean. In addition, an evaluation was done per segment size: The test cases were divided according to the ROI size in pixels and the accuracy of each size range was measured separately (Tables 1–4).

## 5. Results and discussion

The results of the attention methods discussed in section 2 (Figure 3) are given in Tables 1–4. It can be seen that methods based on generating an attention map and combining it with the main branch net branch (Figure 3b-g) gave considerably better accuracy than hard attention methods based on blacking out the background region[3] (Figure 2a). The difference in accuracy is particularly large for the classification of small segments (Tables 1–4) where background information is more important in classification. Merging the attention map with the first layer of the net (Figure 3b–c) gave significantly better results than merging at higher layers (Figure 3d–e). Again, the difference is larger for the classification of small regions. This probably due to the fact that higher layers of the net suffer from a loss of high-resolution information that is relevant in the classification of small objects. Generating several attention maps and merging them with multiple layers of the net (Figure 3f–g) gave the same or worse results than generating a single attention map and merging it with the first layer (Figure 3b–c). Hence, using a single attention map and merging it with the first layer seems to be the best approach in all cases. The method of merging did not seem to have a significant effect, and both addition and multiplication with the attention map had a similar effect, although the addition method seems to give slightly better results in some cases.

Table 1. COCO:mean accuracy per class (equal weights for all classes)

| Net model | Fig 3.c | Fig 3.b | Fig 3.e | Fig 3.d | Fig 3.f | Fig 3.g | Fig 2.a | |
|---|---|---|---|---|---|---|---|---|
| Attention merging layer | First layer | | Third Resnet block | | All Resnet Block | | Image | First layer |
| Merge mode | Addition | Multiplication | Addition | Multiplication | Addition | Multiplication | Background blackout | |
| Region size range in pixels | Mean class accuracy | | | | | | | |
| 0-1000 | 68% | 68% | 54% | 51% | 67% | 66% | 45% | 48% |
| 1000-2000 | 80% | 80% | 75% | 70% | 80% | 79% | 65% | 68% |
| 2000-4000 | 84% | 82% | 80% | 75% | 83% | 83% | 72% | 74% |
| 4000-8000 | 85% | 85% | 83% | 80% | 85% | 85% | 76% | 79% |
| 8000-16000 | 87% | 86% | 85% | 81% | 86% | 86% | 80% | 81% |
| 16000-32000 | 88% | 87% | 86% | 82% | 87% | 87% | 82% | 84% |
| 32000-64000 | 85% | 85% | 85% | 80% | 85% | 85% | 79% | 81% |
| 64000-128000 | 87% | 86% | 85% | 79% | 84% | 85% | 78% | 82% |
| 128000-256000 | 83% | 81% | 79% | 78% | 81% | 83% | 76% | 80% |
| 256000-500000 | 76% | 70% | 70% | 66% | 71% | 73% | 68% | 72% |
| Average accuracy all sizes | 83% | 83% | 77% | 73% | 83% | 82% | 70% | 72% |

Table 2. OpenSurfaces: mean accuracy per class (equal weights for all classes)

| Net model | Fig 3.c | Fig 3.b | Fig 3.e | Fig 3.d | Fig 3.f | Fig 3.g | Fig 2.a | |
|---|---|---|---|---|---|---|---|---|
| Attention merging layer | First layer | | Third Resnet block | | All Resnet Block | | Image | First layer |
| Merge mode | Addition | Multiplication | Addition | Multiplication | Addition | Multiplication | Background blackout | |
| Region size range in pixels | Mean class accuracy | | | | | | | |
| 0-1000 | 25% | 15% | 11% | 9% | 17% | 13% | 9% | 10% |
| 1000-2000 | 49% | 46% | 27% | 32% | 42% | 38% | 40% | 33% |
| 2000-4000 | 49% | 50% | 46% | 35% | 39% | 46% | 33% | 29% |
| 4000-8000 | 57% | 49% | 51% | 40% | 44% | 53% | 39% | 34% |
| 8000-16000 | 53% | 51% | 47% | 41% | 50% | 50% | 41% | 35% |
| 16000-32000 | 56% | 57% | 55% | 46% | 52% | 54% | 43% | 39% |
| 32000-64000 | 59% | 56% | 54% | 49% | 53% | 56% | 48% | 46% |
| 64000-128000 | 66% | 65% | 61% | 60% | 64% | 64% | 54% | 52% |
| 128000-256000 | 72% | 69% | 77% | 65% | 66% | 68% | 62% | 55% |
| 256000-500000 | 71% | 72% | 72% | 65% | 71% | 70% | 68% | 63% |
| 500000-1000000 | 71% | 78% | 71% | 68% | 77% | 79% | 70% | 71% |
| Average accuracy all sizes | 52% | 50% | 46% | 41% | 46% | 49% | 40% | 37% |

### Table 3. COCO, mean accuracy per image

| Net model figure 3 | c | b | e | d | f | g | Fig 2.a | |
|---|---|---|---|---|---|---|---|---|
| Attention merging layer | First layer | | Third Resnet block | | All Resnet Block | | Image | First layer |
| Merge mode | Addition | Multiplication | Addition | Multiplication | Addition | Multiplication | Background blackout | |
| Region size range in pixels | Mean accuracy per image | | | | | | | |
| 0-1000 | 75% | 75% | 58% | 56% | 76% | 75% | 57% | 58% |
| 1000-2000 | 83% | 83% | 77% | 74% | 83% | 82% | 71% | 73% |
| 2000-4000 | 84% | 84% | 80% | 78% | 84% | 84% | 75% | 77% |
| 4000-8000 | 85% | 86% | 83% | 81% | 85% | 85% | 78% | 80% |
| 8000-16000 | 87% | 87% | 85% | 83% | 87% | 87% | 81% | 83% |
| 16000-32000 | 90% | 89% | 87% | 85% | 89% | 88% | 85% | 86% |
| 32000-64000 | 91% | 90% | 89% | 87% | 90% | 89% | 86% | 87% |
| 64000-128000 | 91% | 91% | 90% | 88% | 91% | 90% | 87% | 88% |
| 128000-256000 | 90% | 89% | 87% | 86% | 88% | 88% | 86% | 86% |
| 256000-500000 | 90% | 89% | 85% | 83% | 88% | 87% | 84% | 85% |
| **Average accuracy all sizes** | **82%** | **82%** | **73%** | **71%** | **82%** | **81%** | **70%** | **71%** |

### Table 4. OpenSurfaces, mean accuracy per image

| Net model | Fig 3.c | Fig 3.b | Fig 3.e | Fig 3.d | Fig 3.f | Fig 3.g | Fig 2.a | |
|---|---|---|---|---|---|---|---|---|
| Attention merging layer | First layer | | Third Resnet block | | All Resnet Block | | Image | First layer |
| Merge mode | Addition | Multiplication | Addition | Multiplication | Addition | Multiplication | Background blackout | |
| Region size range in pixels | Mean accuracy per image | | | | | | | |
| 0-1000 | 32% | 23% | 18% | 17% | 21% | 22% | 18% | 20% |
| 1000-2000 | 64% | 67% | 56% | 57% | 60% | 61% | 51% | 51% |
| 2000-4000 | 72% | 71% | 66% | 62% | 66% | 71% | 58% | 60% |
| 4000-8000 | 73% | 72% | 72% | 69% | 69% | 73% | 61% | 63% |
| 8000-16000 | 77% | 75% | 75% | 72% | 74% | 77% | 66% | 66% |
| 16000-32000 | 80% | 79% | 78% | 76% | 76% | 79% | 71% | 70% |
| 32000-64000 | 82% | 81% | 81% | 79% | 79% | 82% | 77% | 75% |
| 64000-128000 | 84% | 84% | 84% | 83% | 84% | 84% | 80% | 79% |
| 128000-256000 | 90% | 90% | 90% | 88% | 90% | 90% | 86% | 87% |
| 256000-500000 | 91% | 90% | 91% | 89% | 92% | 90% | 91% | 90% |
| 500000-1000000 | 89% | 91% | 87% | 88% | 91% | 91% | 88% | 89% |
| **Average accuracy all sizes** | **80%** | **78%** | **78%** | **75%** | **77%** | **79%** | **72%** | **72%** |

# Supporting materials

Code and weights for the region based classification nets can be found in the links below:

Object classification trained on COCO dataset:

https://github.com/sagieppel/Classification-of-object-in-a-specific-image-region-using-a-convolutional-neural-net-with-ROI-mask-a

Material classification trained on OpenSurfaces dataset:

https://github.com/sagieppel/Classification-of-the-material-given-region-of-an-image-using-a-convolutional-neural-net-with-attent